\documentclass{article}

\usepackage{arxiv}

\usepackage[utf8]{inputenc} 
\usepackage[T1]{fontenc}    
\usepackage{hyperref}       
\usepackage{url}            
\usepackage{booktabs}       
\usepackage{amsfonts,amsmath,amssymb}       
\usepackage{nicefrac}       
\usepackage{microtype}      
\usepackage{lipsum}		
\usepackage{graphicx}
\usepackage{doi}
\usepackage{subfigure}
\usepackage{algorithm}
\usepackage{algorithmic}
\usepackage{multirow}
\usepackage{float}

\title{Accurate Fine-grained Layout Analysis for the Historical Tibetan Document Based on the Instance Segmentation\thanks{This study was funded by the National Natural Science Foundation of China (No.61772430, No.61375029), program for Leading Talent of State Ethnic Affairs Commission, program for Innovative Research Team of SEAC ([2018]98), the Gansu Provincial first-class discipline program of Northwest Minzu University (No.11080305), Natural Science Foundation of Gansu Province of China (No. 21JR1RA195), and the Postgraduate Support Programs of Northwest Minzu University's Fundamental Research Funds for the Central Universities (Ymx2021002). }\thanks{This work has been submitted to the IEEE for possible publication. Copyright may be transferred without notice, after which this version may no longer be accessible.}}

\author{{\hspace{1mm}Penghai Zhao}\\
	Key Laboratory of China's Ethnic Languages\\
	and Information Technology of Ministry of Education\\
	Northwest Minzu University\\
	Lanzhou, 730000 China \\
	\And
{\hspace{1mm}Weilan Wang}\thanks{Corresponding author: Weilan Wang (e-mail: wangweilan@xbmu.edu.cn).}\\
Key Laboratory of China's Ethnic Languages\\
and Information Technology of Ministry of Education\\
Northwest Minzu University\\
Lanzhou, 730000 China \\
\And
{\hspace{1mm}Zhengqi Cai}\\
School of Mathematics and Computer Science\\
Northwest Minzu University\\
Lanzhou, 730000 China \\
\And
{\hspace{1mm}Guowei Zhang}\\
Key Laboratory of China's Ethnic Languages\\
and Information Technology of Ministry of Education\\
Northwest Minzu University\\
Lanzhou, 730000 China \\
\And
{\hspace{1mm}Yuqi Lu}\\
Key Laboratory of China's Ethnic Languages\\
and Information Technology of Ministry of Education\\
Northwest Minzu University\\
Lanzhou, 730000 China \\
}

\renewcommand{\shorttitle}{Accurate Fine-grained Layout Analysis for the Historical Tibetan Document}

\hypersetup{
pdftitle={Accurate Fine-grained Layout Analysis for the Historical Tibetan Document Based on the Instance Segmentation},
pdfsubject={q-bio.NC, q-bio.QM},
pdfauthor={David S.~Hippocampus, Elias D.~Striatum},
pdfkeywords={First keyword, Second keyword, More},
}

\begin{document}
\maketitle

\begin{abstract}
Accurate layout analysis without subsequent text-line segmentation remains an ongoing challenge, especially when facing the Kangyur, a kind of historical Tibetan document featuring considerable touching components and mottled background. Aiming at identifying different regions in document images, layout analysis is indispensable for subsequent procedures such as character recognition. However, there was only a little research being carried out to perform line-level layout analysis which failed to deal with the Kangyur. To obtain the optimal results, a fine-grained sub-line level layout analysis approach is presented. Firstly, we introduced an accelerated method to build the dataset which is dynamic and reliable. Secondly, enhancement had been made to the SOLOv2 according to the characteristics of the Kangyur. Then, we fed the enhanced SOLOv2 with the prepared annotation file during the training phase. Once the network is trained, instances of the text line, sentence, and titles can be segmented and identified during the inference stage. The experimental results show that the proposed method delivers a decent 72.7\% AP on our dataset. In general, this preliminary research provides insights into the fine-grained sub-line level layout analysis and testifies the SOLOv2-based approaches. We also believe that the proposed methods can be adopted on other language documents with various layouts.
\end{abstract}


\section{Introduction}
\label{sec:1}
Historical Tibetan documents are regarded as one of the most desirable materials for the study of ancient Tibetan culture. To protect these invaluable historical documents and make rapid retrieval possible, archiving the corresponding digital document images has been taken as a practical solution. However, further digitalization is limited by the voluminous historical documents and chronically understaffed protection organizations. If these archived document images are automatically analyzed and recognized into machine-encoded text, specialists will no longer need to type manually and thus have more time for the better protection of the historical documents.

\begin{figure}[h]
\centering
\includegraphics[width=\columnwidth]{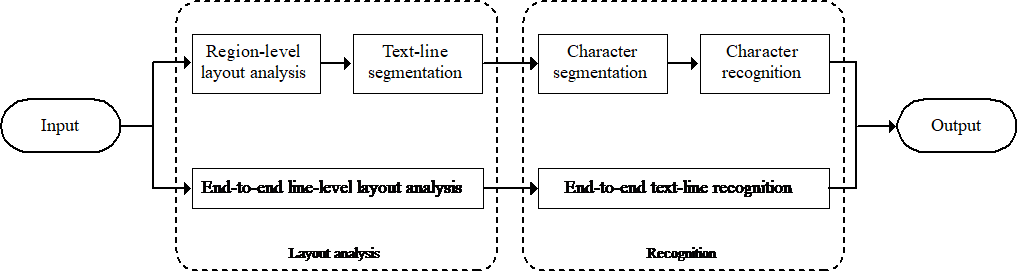}
\caption{Normal procedures of DAR include traditional approach (top) and deep learning-based approach (bottom)}
\label{fig1}
\end{figure}

Document Analysis and Recognition (DAR) system is designed to automatically extract information from the document images and transform them into a digital symbolic representation. Fig. \ref{fig1} presents an overview of the existing DAR system. As can be seen in Fig. \ref{fig1}, layout analysis is an important component in the DAR system which aims at the detection of various regions to be recognized including text regions, image regions, and individual sub text-line regions in historical Tibetan document images. For text and sub-line regions, there will be subsequent recognition algorithms \cite{1,2} converting them into the encoded text.

\begin{figure}[h]
	\centering
	\includegraphics[width=\columnwidth]{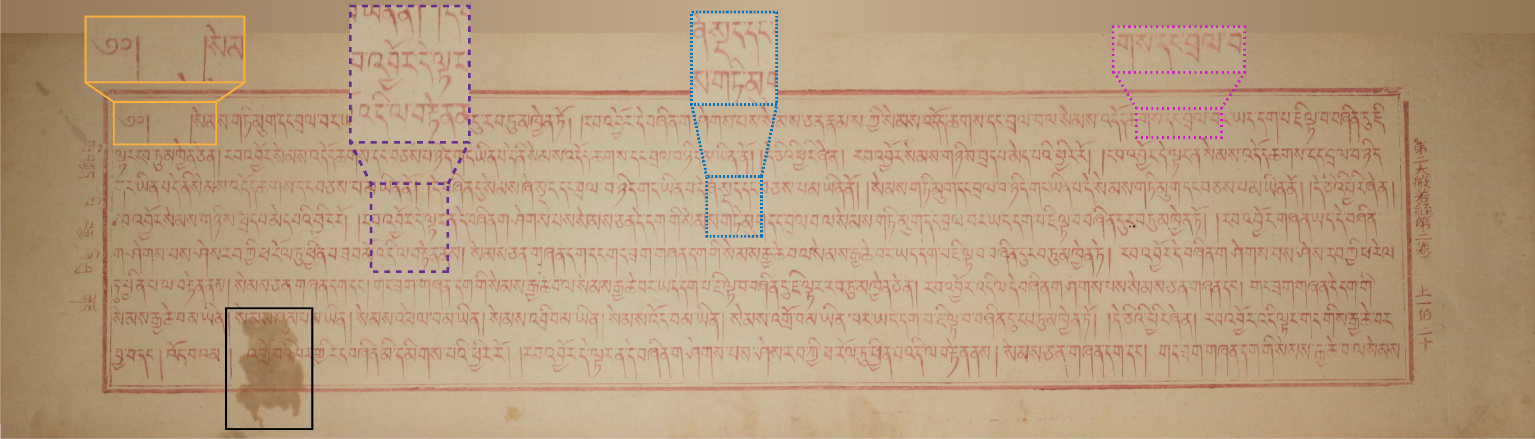}
	\caption{Three features of the historical Tibetan document image: stains (black box), closed or touching strokes (purple and blue box), faded strokes (pink box), and excessive space between sentences (orange box)}
	\label{fig2}
\end{figure}
Most document layout analysis approaches utilized semantic segmentation neural networks \cite{3,4,5} to divide document images into various zones. Such networks, however, have failed to distinguish different text lines in text blocks, and therefore extra text-line segmentation methods were required to decompose the regions into text line fragments containing a sequence of characters. This would lead to a highly coupled system which may impede the improvement of ultimate performance. For example, if each method used in the traditional approach of Fig. \ref{fig1} is assumed to deliver a strong performance of 90\% accuracy, the final accuracy could be only 65.6\% due to the cumulative error.

To date, far too little attention has been paid to line-level layout analysis which can separate distinct text lines. A few existing line-level layout analysis research treated layout analysis as an instance segmentation task and employed the instance segmentation network to identify different text lines, such as Mask R-CNN \cite{6}. Mask R-CNN is capable of extracting individual text lines from the image and is proved to be effective on certain datasets. Unfortunately, the majority of instance segmentation networks are unable to process the Kangyur images, a typical kind of historical Tibetan document images, in which various sentence lengths and spaces, close or touching strokes, and a considerable number of stains can be found (see Fig. \ref{fig2}). In addition, the practical applications of the deep learning-based approach have been hampered by the lack of accurate line-level annotations. The construction of the dataset is still facing challenges.

This paper proposes enhancing SOLOv2 \cite{7} by improving the backbone in terms of the characteristics of the Kangyur images and utilizing it for sub-line level layout analysis. First, accelerated fine-grained line-level annotations are semi-automatically generated by adopting the traditional text-line segmentation method and manual revisions. Ten classes are considered based on the object type and location which are line $i$, $i\in[1,8]$, `ltitle', and `rtitle'. Then, feeding the improved SOLOv2 with generated data during the training phase. Third, utilizing the trained network to obtain the final layout analysis results. We do not employ any pre-processing and post-processing strategies. The major contributions are:
\begin{enumerate}
\item To the best of our knowledge, this paper appears to be the first study adopting the SOLOv2-based network to perform accurate fine-grained sub-line level layout analysis.
\item This study provides insights into the accelerated construction of the dataset which is dynamic and accurate.
\item Our proposed method is effective both from quantitative results and visual effects.
\end{enumerate}

The remaining part of the paper proceeds as follows: The second section is concerned with the related work. Section Three details the method and reports the evaluation metrics. Dataset construction, quantitative results, and visualization results are provided in Section Four. Finally, the conclusion and a brief overview of future work are presented in Section Five.

\section{Related work}
\label{sec:2}
\textbf{Layout Analysis}. As an important procedure of document analysis and recognition, layout analysis has attracted extensive interest over the past decades. Numerous methods have been proposed to parse the layout of different documents and they can be categorized into two major classes: traditional methods and deep learning-based methods.

Traditional methods were widely used due to their interpretability and lower-level time complexity. F. Zirari \textit{et al}. proposed modeling the image by using a graph to establish the close connexity and relations according to pixels intensity-based homogeneity criterion which can efficiently separate graphical and textual parts of the document images \cite{8}. Jewoong Ryu presented a connected components analysis-based method that achieved state-of-the-art performance on the Latin-based and Chinese scripts in 2014 \cite{9}. Abedelkadir Asi exploited texture-based filters in combination with Markov Random Fields (MRF) and provided a framework to identify different text regions which obtain promising results on a subset of the public dataset \cite{10}. However, these methods rely heavily on heuristic rules and require a number of parameters to improve the performance. When the layout of a document is relatively complex, these methods may fail to deliver the optimal results.

Most recent research in layout analysis has emphasized the use of deep neural networks, especially the semantic segmentation network. Kai Chen \textit{et al}. considered the layout analysis as a pixel classification task and employed three-level convolutional autoencoder architecture to learn feature representation from pixels. The experiments carried out on three public datasets had demonstrated the effectiveness of the proposed method \cite{11}. Yue xu \textit{et al}. fed FCN with pixel-wise labels and utilized it to classify pixels into four classes \cite{12}. More than 99\% pixel-level accuracy demonstrated the superiority of the FCN based approach. In the same year, more improved FCN-based methods had been proposed by Xiao Yang \textit{et al}. \cite{13} and Dafang He \textit{et al}. \cite{14}, and both of them obtained promising results.

Despite the acceptable segmentation results of semantic segmentation network-based layout analysis, the obtained regions are still needed to be further split into text lines to meet the requirements of subsequent recognition algorithms. Projection-based line segmentations are one of the most commonly used methods and can be applied to most printed document images. For documents with a more complex layout, advanced approaches have been proposed. Ma \textit{et al}. used block projection to divide images into text, line, and frames, and then utilized a graph model-based text line segmentation method to address the problem of touching strokes \cite{15}. Wang \textit{et al}. pointed out that allocating connected components rather than pixels to the text line might reduce the noise and thus achieved finer performance \cite{16}. A text-line segmentation approach exploiting local baselines and connected components was presented by Pengfei Hu which obtained decent segmentation accuracy on the historical document images \cite{17}.

There are relatively few studies in the area of instance segmentation-based layout analysis. Xiao-Hui Li presented a U-shaped segmentation network called LPN to obtain region-level layout analysis results. With employing an additional watershed transformation, these semantic regions are finally converted to the instance aware regions including text lines, figures, and tables, etc. \cite{18}. Being different from semantic segmentation, end-to-end instance segmentation separates different objects of the same label. Abdullah Almutairi conducted a region-level layout analysis by employing the instance segmentation network Mask R-CNN which is language-agnostic and can deconstruct newspaper images into various parts \cite{19}. Abhishek Prusty \textit{et al}. adopted Mask R-CNN for automatic, instance-level spatial layout parsing of historical Indic manuscripts images. In his major study, Prusty demonstrated the effectiveness of the Mask R-CNN which reached 64.76\% of AP50 on the Indiscapes dataset \cite{20}. However, Mask R-CNN cannot be applied to the historical document images directly because of the frequent occurrence of segmentation mistakes.

\textbf{Dataset}. Accurate annotations are indispensable for pixel-wise layout analysis and the process of annotating is always extremely exhausting. Over the past ten years, only a few layout analysis datasets were released including the widely used PubLayNet \cite{21} and DIVA-hisDB \cite{22}. With a total number of more than 360 thousand document images at the polygon-level, the PubLayNet dataset was generated by automatically matching the XML representations and the content of PDF from PubMed CentralTM. The annotation of the DIVA-hisDB was generated by adopting the approach mentioned in \cite{23}. Its annotated region's edge shows much more precision compared to the PubLayNet, but likewise failed to establish the distinction of individual text lines. Collectively, these studies outline an essential role in the construction of the dataset. But for now, there is still no available method that can be directly adopted to address the lack of an accurate instance-aware dataset for the historical Tibetan document layout analysis.

\section{Methods}
\label{sec:3}
\subsection{Motivation}
\begin{figure}[h]
	\centering
	\includegraphics[width=.8\columnwidth]{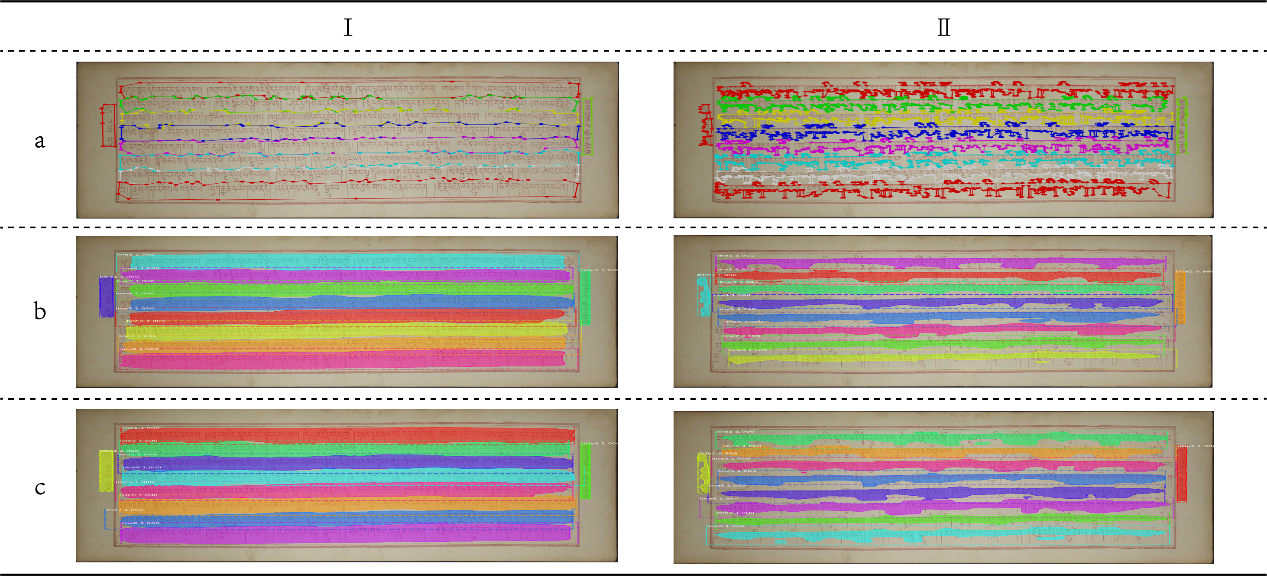}
	\caption{Results of different Mask R-CNN and annotations: (a) annotations; (b) results using vanilla Mask R-CNN; (c) results using improved Mask R-CNN}
	\label{fig3}
\end{figure}

Mask R-CNN was adopted to parse the layout of historical Tibetan images in response to the views of Prusty A's studies \cite{20}. As can be seen in the left column of Fig. \ref{fig3} (a), we used the Labelme tool \cite{24} to roughly annotate ten images and trained the network to try to fully exploit the potential of Mask R-CNN (even overfitting could happen).

The results of our experiment are similar to Prusty's in which segmentation errors might be found on both sides of text lines (see the left column of Fig.\ref{fig3} (b). A possible explanation for these errors may be the restriction of the ground truth region mask which is typically limited to 28×28. Such restriction causes considerable information loss, and therefore we upsampled the mask head to 112$\times$112 by employing two additional transposed convolutional layers to try to recover the high-resolution representation. It can be seen from the left column of Fig. \ref{fig3} (c) that a larger mask head is favorable to the avoidance of segmentation errors on both ends.

However, the improved Mask R-CNN training with the roughly annotated data failed to produce acceptable results, probably due to adjacent or overlapped text lines. We assumed a more accurate fine-grained annotation may improve the results, and thus prepared the annotated data depicted in the right column of Fig.\ref{fig3} (a). Surprisingly, more accurate annotations did not lead to a better result. Neither vanilla nor improved Mask R-CNN obtained satisfactory results (see right column of Fig.\ref{fig3} (b) (c)).

Nevertheless, it seems that the enlarged mask head is beneficial for segmentation results no matter what type of data is used. Unfortunately, we can no longer scale the mask head up due to the limitation of the GPU. Therefore, we would like to employ an instance segmentation network without the size restriction of the mask head. By investigating the COCO dataset, Wang X suggested that the fundamental differences between object instances could be their center locations and sizes and proposed two promising instance segmentation networks named SOLO \cite{25} and SOLOv2 \cite{7}. Compared with Mask R-CNN, these networks remove the restriction on the mask head and may generate higher-quality masks.

\subsection{Network architecture}
Individual objects are identified by SOLO according to their locations and sizes. As a result, SOLO gains a natural advantage from such strategies when processing document images containing different regions with various shapes and locations. To segment objects by location, the input image is divided into $S \times S$ grids. If the center of the object locates at ${grid}(i, j)$, the grid will be used for categorizing and segmenting by two parallel branches. The output space of category branch and mask branch are $S \times S \times C$ and $H \times W \times S^{2}$ respectively, where $C$ is the number of classes and $H, W$ are the height and width of the input image. The $K_{t h}$ output channel of mask branch can be one-to-one corresponded to ${grid}(i, j)$ according to $i \times S+j$ (with $i, j$ zero-based).

However, SOLO suffers from certain drawbacks which are further addressed in SOLOv2. Given that most images contain sparse objects, the channels of SOLO's mask branch could be somewhat redundant. By integrating dynamic convolution and Matrix NMS, SOLOv2 can separately learn pyramid features $F$ and convolution kernel $G$ which can improve the details on object boundaries. SOLOv2 produces finer performance on the COCO dataset and we tried to enhance it to generate higher-quality masks of Tibetan text lines.

\textbf{Backbone}. Convolution neural network, usually called backbone in the multi-task network, is used for automatically extracting features from the image. The adopted backbone of SOLOv2 is composed of the FPN \cite{26} which has two pathways. The bottom-up pathway uses ResNet \cite{27} to extract features at different levels. High-level features are of great semantic value but contain less information of small objects. Based on these semantic rich features, the top-down pathway is utilized to better preserves the information of small objects and reconstructs high-resolution features by a series of upsampling layers. Thus, FPN is effective for extracting features of both large and small objects, which are entire text lines and short sentences in the historical Tibetan document images.

Considering the higher resolution of the historical Tibetan document images, specific improvements have been made to the backbone of SOLOv2. Similar structure of the high-resolution feature pyramid network (HRFPN) \cite{28} was designed to augment the high-resolution representation of the backbone. In addition, we replace the ResNet in the bottom-up pathway with ResNeXt \cite{29} which is supposed to demonstrate a stronger performance than the original one.

\begin{figure}[h]
	\centering
	\includegraphics[width=.6\columnwidth]{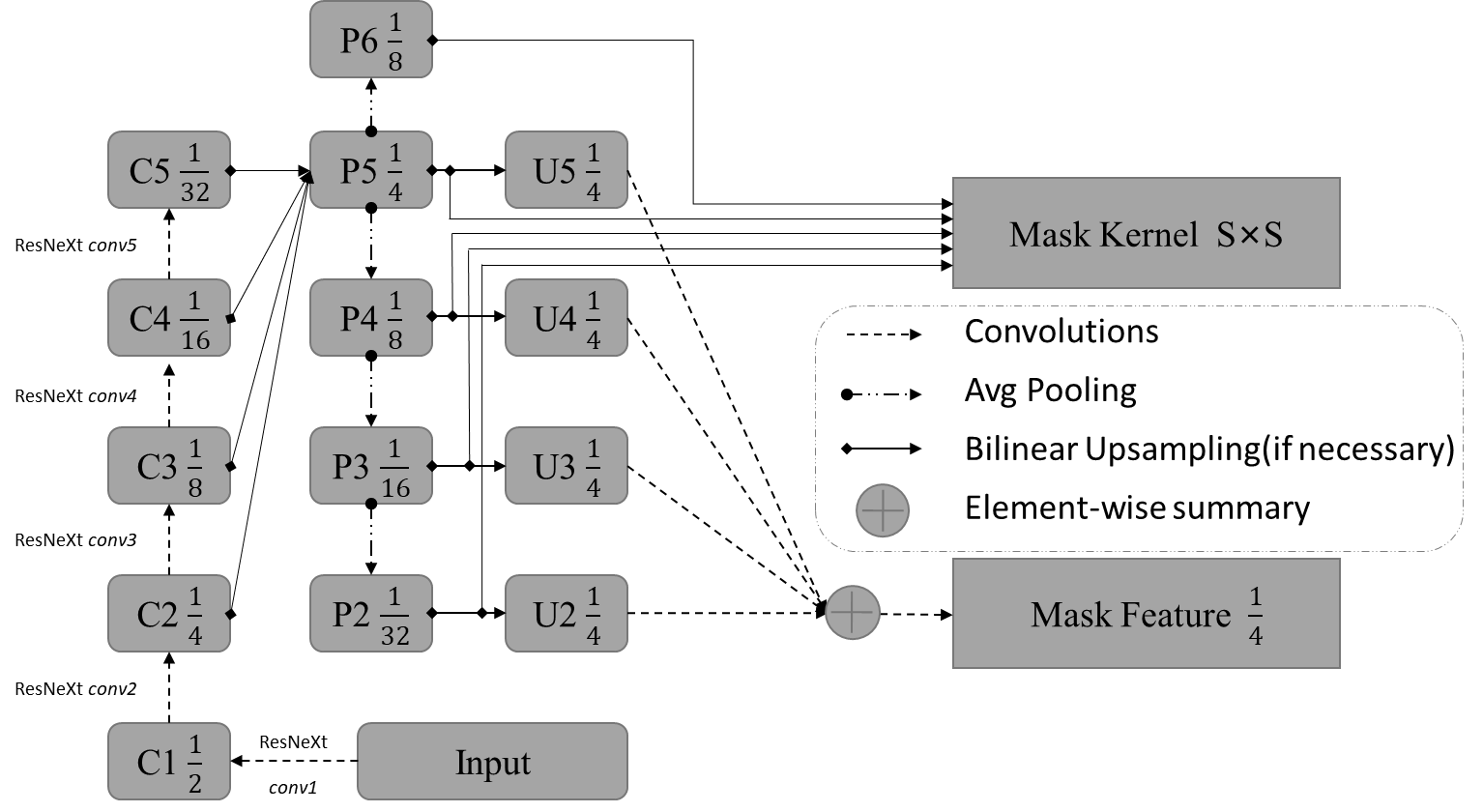}
	\caption{The proposed backbone: fraction denotes the feature map size comparing to the input image}
	\label{fig4}
\end{figure}

Fig. \ref{fig4} details the proposed backbone. C1, C2, $\ldots$, C5 corresponding to the output feature maps at different stages of the ResNeXt. The low-resolution feature maps C3, C4, C5 will be upscaled to the same size as C2 using bilinear interpolation. We feed the topdown pathway of HRFPN with the concatenation of feature maps from C2 to C5. Pyramid feature maps {P}2, {P}3, {P}4 and {P}6 are down sampled to highlight the average presence of the feature from the corresponding upper layer. For P2 to P5, each of them is upscaled by repeated convolutions and bilinear interpolations until it reaches $1 / 4$ scale of the input image. An element-wise summation is performed on these unified mask feature maps, and a $1 \times 1$ convolution is used to generate the final output mask feature $F \in \mathbb{R}^{h \times w \times E}$. For P2 to {P}6, all of them will be resized to ${S} \times {S}({S}=12,16,24,36,40$ respectively) as inputs of the mask kernel branch.

\begin{figure}[h]
	\centering
	\includegraphics[width=\columnwidth]{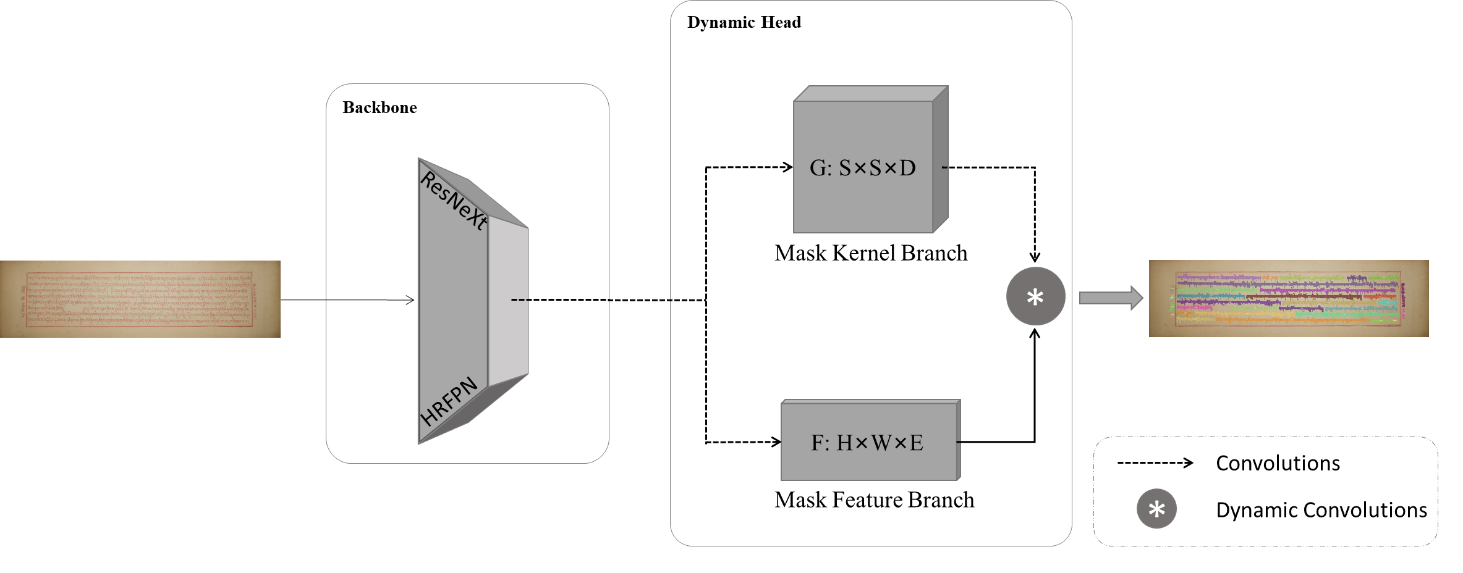}
	\caption{ The improved SOLOv2 with the use of HRFPN and ResNeXt}
	\label{fig5}
\end{figure}

\textbf{Dynamic Head}. The dynamic head consists of the mask kernel branch and mask feature branch. Mask kernel branch is designed to learn the dynamic convolution kernel where the input feature $F$ with the size of $h\times w\times E$ and the output space is $S\times S\times E$ ($h$ and $w$ are height and width of output at each stage of FPN, $E$ correspond to input channels). Mask feature aims at segmenting the objects. Its input is the unified feature with the size of $(H/4)\times(W/4)\times E$ and the output space is $H\times W\times E$ ($H$ and $W$ are the height and width of the input image). The final results are further produced by the kernels $G$ dynamic convolving with the output $F$ of the mask feature branch \cite{30,31}. In addition to the decrease in parameters, the generation of the kernel is dynamically conditioned on the input which endows SOLOv2 with more adaptability and flexibility.

The architecture of the modified network is illustrated in Fig. \ref{fig5}. Hyperparameters and other model details remain as consistent as possible as \cite{7}.

\subsection{Evaluation metrics}
We adopted the widely used Average Precision (AP) to access the performance of several approaches \cite{32}. AP, the primary metric, is averaged over 10 Intersection over Union (IoU) thresholds from 0.5 to 0.95. Additionally, we used AP$_{50}$ and AP$_{75}$ which stand for AP at IoU$=$0.50, 0.75 respectively. For these metric values, the higher, the better.

\section{Results}
\label{sec:4}
The Peking edition of Kangyur (PeK) is now collected in the National Library of Mongolia and has been digitalized by the Asian Classics Input Project (ACIP) and Yuishoji Buddhist Cultural Exchange Research Institute (YBCERI) since 2008. We carefully selected 212 samples from thousands of Kangyur images to build the Peking Kangyur Layout Analysis Dataset (PKLAD). The PKLAD contains 144 training samples and 34 validation and test samples and each of them has precisely 8 text lines and 2 titles.

\begin{figure}[h]
	\centering
	\includegraphics[width=.5\columnwidth]{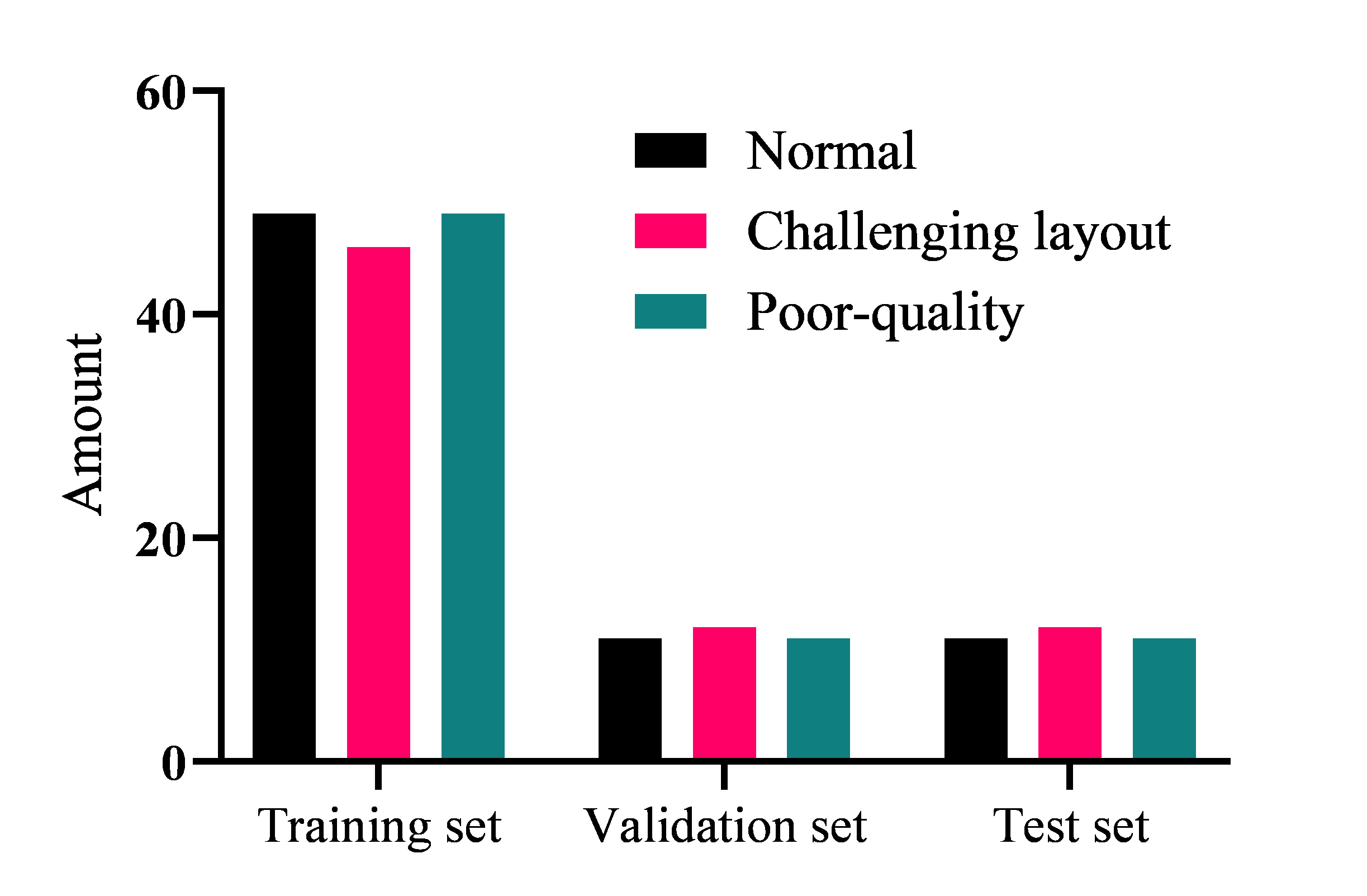}
	\caption{Dataset composition}
	\label{fig6}
\end{figure}

\begin{figure}[h]
\centering
\subfigure[]{
\begin{minipage}{0.2\columnwidth}
\includegraphics[width=\columnwidth]{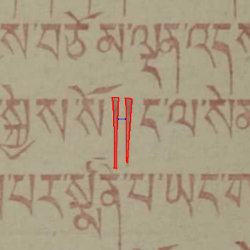}
\end{minipage}}
\hspace{1cm}
\subfigure[]{
	\begin{minipage}{0.2\columnwidth}
		\includegraphics[width=\columnwidth]{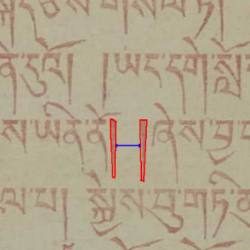}
\end{minipage}}
\hspace{1cm}
\subfigure[]{
	\begin{minipage}{0.2\columnwidth}
		\includegraphics[width=\columnwidth]{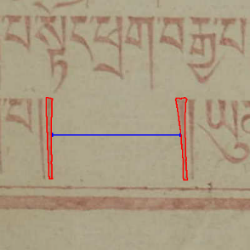}
\end{minipage}}
\caption{Various spaces between sentences: `|' with the red contour can be regarded as a period in English. (a) represents a narrow space with average pixels of 10, (b) represents an average pixel distance of 40, whereas space in (c) can reach up to 200px or even more.}
\label{fig7}
\end{figure}

Each set of the PKLAD may be divided into three categories including normal type, challenging layout, and poor quality in terms of different problems they could cause. Fig. \ref{fig6} presents the composition of the PKLAD. The challenging layout category is those images with skewed or warped pages, extreme close or distant text lines, and insufficient or excessive space between sentences (see Fig. \ref{fig7}). Images featuring non-uniform illumination, mottled background, and faded strokes belong to the poor-quality category.

\textbf{Fine-grained annotations}. Annotations of the PeK images are fine-grained. The definition of fine-grained is two-fold. (i) The class is fine-grained. We use line1, line2, \ldots, line8, and sub-line level labels to separate different text regions rather than treating all text lines as the same text region. (ii) The region is fine-grained. Being different from other common polygon annotations, fine-grained annotations provide more precise details, especially for boundaries of the text regions.

\textbf{Annotation challenges}. Fine-grained annotations enjoy several advantages including better visual effects and higher accuracy. However, the procedure of labeling the PeK images in a fine-grained manner is full of difficulties. Firstly, the time frame of annotation can be unacceptable. Due to the huge resolution of these document images (around 5500$\times$1500 resolution), the fine-grained annotation of only one single image could stretch over 2 hours. Secondly, errors are inevitable. Annotators have different senses of the character boundary, and this would lead to the inconsistencies of the annotation especially when the boundary is blurred. Furthermore, all humans make mistakes. Even with a subsequent revision procedure, it is still impossible to correct all the mistakes.

\begin{figure}[h]
	\centering
	\includegraphics[width=\columnwidth]{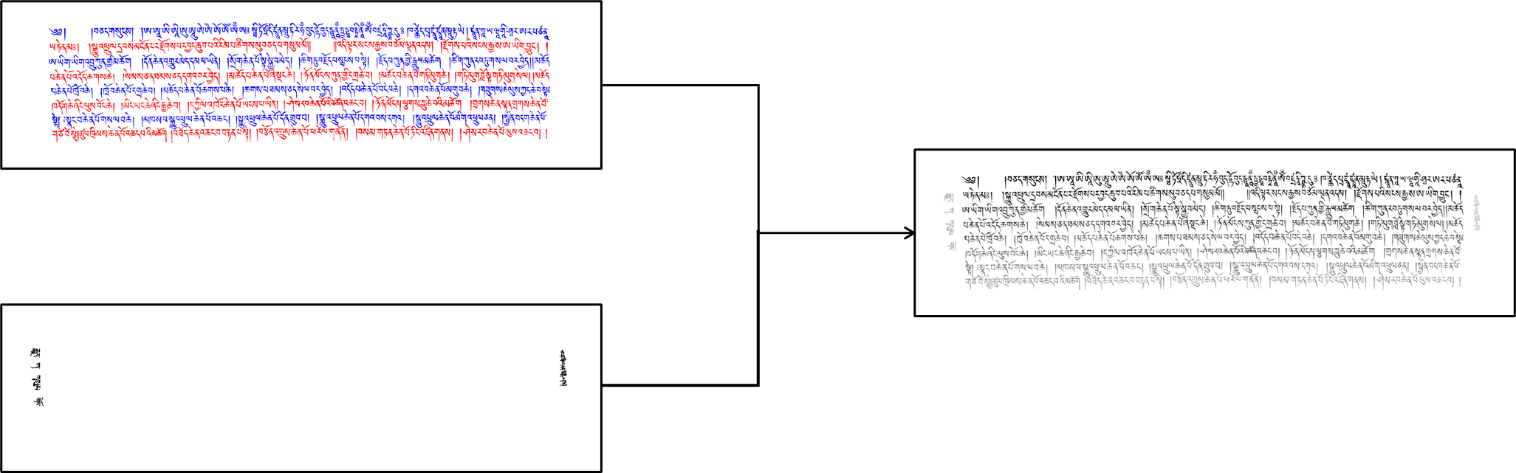}
	\caption{Obtain the intensity image using bitwise AND operation}
	\label{fig8}
\end{figure}

\textbf{Semi-automatic annotation method}. As manual annotation faces various challenges, we proposed a semi-automatic annotation approach which consists of three major steps.

(1) Obtain the intensity image of the text-line segmentation result. To obtain the resulted intensity image, the first step is to segment and extract different text lines from the image. Any text-line segmentation approaches are welcome. Here, we borrowed the ideas from the previously proposed research \cite{16,17} to decompose the image into several text lines. Due to the extra and unique titles of the Kangyur, we also manually labeled them with the use of Adobe Photoshop. Then, we combined the text-line segmentation results with the corresponding titles, and set the pixel value of $i_{th}$ text line to $20\times i,i\in[1,8]$. The pixel value of the left title and right title is 180 and 200 respectively. There are ten classes in total being considered in the PKLAD. The resulted intensity image is depicted in the right column of Fig. \ref{fig8}.

\begin{figure}[h]
	\centering
	\subfigure[6-0]{
		\begin{minipage}{0.15\columnwidth}
			\includegraphics[width=\columnwidth]{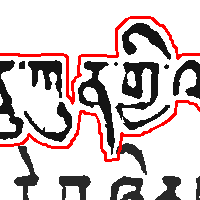}
	\end{minipage}}
	\hspace{1cm}
	\subfigure[6-2]{
		\begin{minipage}{0.15\columnwidth}
			\includegraphics[width=\columnwidth]{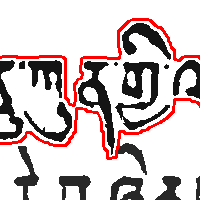}
	\end{minipage}}
	\hspace{1cm}
	\subfigure[6-4]{
		\begin{minipage}{0.15\columnwidth}
			\includegraphics[width=\columnwidth]{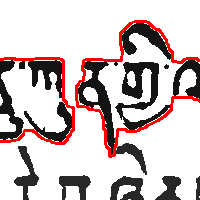}
	\end{minipage}}
\hspace{1cm}
\subfigure[6-6]{
	\begin{minipage}{0.15\columnwidth}
		\includegraphics[width=\columnwidth]{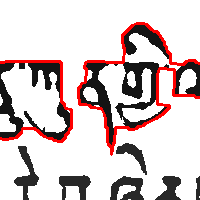}
\end{minipage}}\\
	\subfigure[10-4]{
	\begin{minipage}{0.15\columnwidth}
		\includegraphics[width=\columnwidth]{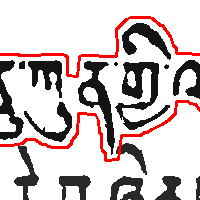}
\end{minipage}}
\hspace{1cm}
\subfigure[10-6]{
	\begin{minipage}{0.15\columnwidth}
		\includegraphics[width=\columnwidth]{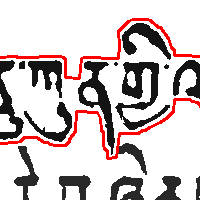}
\end{minipage}}
\hspace{1cm}
\subfigure[10-8]{
	\begin{minipage}{0.15\columnwidth}
		\includegraphics[width=\columnwidth]{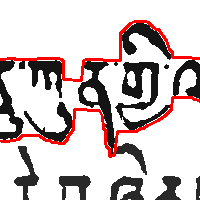}
\end{minipage}}
\hspace{1cm}
\subfigure[10-10]{
	\begin{minipage}{0.15\columnwidth}
		\includegraphics[width=\columnwidth]{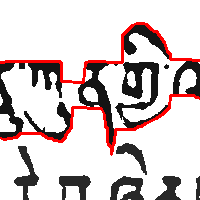}
\end{minipage}}
	\caption{Eight types of contours are considered: i.e., (b) represents contours generated by 6 times dilation and 2 times erosion}
	\label{fig9}
\end{figure}

(2) Extract contour of the regions. In this step, we first resize them to the fixed 1504×1504 resolution and separate the resulted intensity image into ten images. Each of these images contains different objects of the same class. Then, these images are binarized, and optionally the morphological opening operation is performed to removing tiny objects such as noise. After that, we apply p times dilation and q times erosion on them. As a result, one or more connected components (CCs) with $(p-q)$ pixels of boundary expansion are obtained. Such a procedure minimizes the number of the CCs and prevents strokes touching caused by dilation in the meantime. A $3\times3$ rectangular structuring element is used for all morphological operations. Finally, we used the algorithm \cite{33} to retrieve all contours from these binary images and adopted the Teh-Chin chain approximation algorithm \cite{34} to compress the contours for the purpose of reducing loading time and saving memory. Fig. \ref{fig9} depicts various types of the Ground Truth (GT) contour.

We provide the pseudocode of step (2) contour extraction as below:
\begin{algorithm}[h]
\caption{Find Contours Algorithm}
\hspace*{0.02in} {\bf Input:}
Intensity Image IMG\\
\hspace*{0.02in} {\bf Output:}
Nested list RST
\begin{algorithmic}[1]
\STATE \textbf{Initialize}: Set i=0
\STATE IMG $\leftarrow$ Resize(IMG)
\STATE [line1, line2, \ldots, line8, ltitle, rtitle] $\leftarrow$ split(IMG) $\backslash\backslash$ Each object in the list is an image containing instances of one class.
\FOR{obj in [line1, line2, \ldots, line8, ltitle, rtitle]}
\STATE  obj = bin(obj) $\backslash\backslash$ Binarize the image.
\STATE  region $\leftarrow$ OpeningDilateErode(obj, p, q) $\backslash\backslash$ Perform corresponding morphological operations on the image.
\STATE CCs $\leftarrow$ getConnectedComponents(region)
\STATE  contour $\leftarrow$ FindContour(CC\_s) $\backslash\backslash$ Adopting Suzuki's method to find contour(s).
\STATE  contour $\leftarrow$ SimplifyContour(contours) $\backslash\backslash$ Adopting C-H Teh's method to simplify contour(s).
\STATE RST[i] $\leftarrow$ contour
\STATE i $\leftarrow$ i+1
\ENDFOR
\end{algorithmic}
\end{algorithm}

(3) Generate the annotation file. There are multiple ways to generate the required annotation file. For example, one plausible solution is writing the class of the CCs, coordinates of continuous points along the contour, and other related information to JSON file in the labelme required format, and use labelme2coco \cite{35} tool to convert it into COCO format JSON file. From top to bottom of the PeK images, the class of text line is line1, line2, \ldots, line8. The class of left and right titles is set to `ltitle' and `rtitle' respectively. If a certain text line contains more than one CC (probably caused by the excessive space between sentences), the group\_id of these CCs will be set to different values.

Thus far, a fine-grained sub-line level annotation file can be generated. Besides, we do not implement any data augmentation strategy (such as geometric transformations, noise injection, or random stains), despite the fact that this may improve the robustness of the method.

\textbf{Advantages of the proposed annotation method}. The main advantages of the proposed methods can be discussed under three headings, which are: (1) Accurate. The boundary of the CCs is automatically retrieved based on the binary images. This may help to reduce man-made errors. (2) Dynamic. We can simply adjust contours by setting different values of $p$ and $q$. It allows us to achieve optimal results on different kinds of the dataset. (3) Time-saving. The required annotation file is generated within only a few minutes.

\subsection{Experimental details}
We build the network using Pytorch, OpenCV, mmdection \cite{36}, and other related tools. The experiments were conducted on the Ubuntu server equipped with multiple NVIDIA K80 cards.

Training. The ResNext used in the backbone was pre-trained on the ImageNet, a large labeled dataset of real-world images, which is proved to be beneficial for network convergence and stabilization. Unless otherwise specified, all deep neural networks were trained using SGD optimizer (with the learning rate$=$0.01, momentum$=$0.9, weight$-$decay$=$0.00001). The batch size and training epoch are set to 2 and 100 respectively. The loss function is described below:
\begin{equation}
L=L_{\text {cate }}+\lambda L_{\text {mask }}
\end{equation}
where category loss $L_{\text {cate }}$ is computed according to the focal loss \cite{37} and mask loss $L_{\text {mask }}$ is Dice Loss \cite{38}. In addition, coefficient $\lambda$ is set to 3.

\textbf{Inference}. The proposed method works fine with the default hyperparameter values of SOLOv2. However, we still changed a few sets of them which are listed in Table \ref{tab1}. Note that the size of the image and the contour type of annotation during inference should remain the same as the data during the training phase.

\begin{table}[h]
\centering
\caption{HYPERPARAMETERS CONFIGURATION}
\label{tab1}
\begin{tabular}{cc}
\hline\hline
Hyperparameter Description&	Value\\\hline
Numbers of RoIs before non-maximal suppression (NMS)&	500 $\rightarrow$ 800\\
Maximum number of objects per image&	100 $\rightarrow$ 500\\
The threshold of visualization of the masks&	0.3 $\rightarrow$ 0.25\\\hline\hline
\end{tabular}
\end{table}

\subsection{Influence of different data}
In this section, we carry out a series of experiments to fully assess the impact of different data. Two major perspectives of training data, input scale and contour type, will be discussed in detail as follows:

Input scale. The average resolution of the PeK images is around 5000$\times$1600, which can consume considerable memory and time for training and inference, or even fail to run on some high-performance computers. Therefore, we have to scale down these images to the appropriate size using bilinear interpolation. The input scale experiment is designed to find the optimal size of the input image which adopts SOLOv2 with a ResNet-50 and FPN backbone (Res-50-FPN) and the `10-8' contour type. Table \ref{tab2} represents the average time per epoch (Avg\_T), memory consumption, and quantitative metrics of varied resolutions. What can be seen in Tab.\ref{tab2} is the steady growth of AP when the resolution becomes larger. Moreover, the nearest aspect ratio compared to the original image leads to better results.

\begin{table}[h]
\centering
\caption{ INFLUENCE OF VARIED INPUT RESOLUTION OF SOLOV2 (USING 10-8 CONTOUR TYPE)}
\label{tab2}
\begin{tabular}{cccccc}
\hline\hline
Resolution&	Avg\_T&	Memory&	AP&	AP$_{50}$&	AP$_{75}$\\\hline
1024$\times$1024&	2.6s&	5500MB&	0.476&	0.859&	0.571\\
1504$\times$1504&	5.4s&	11300MB	&0.571&	0.877&	0.752\\
2496$\times$800&	4.9s&	10100MB&	0.641&	0.908&	0.793\\\hline\hline
\end{tabular}
\end{table}

\begin{figure}[h]
	\centering
	\includegraphics[width=.5\columnwidth]{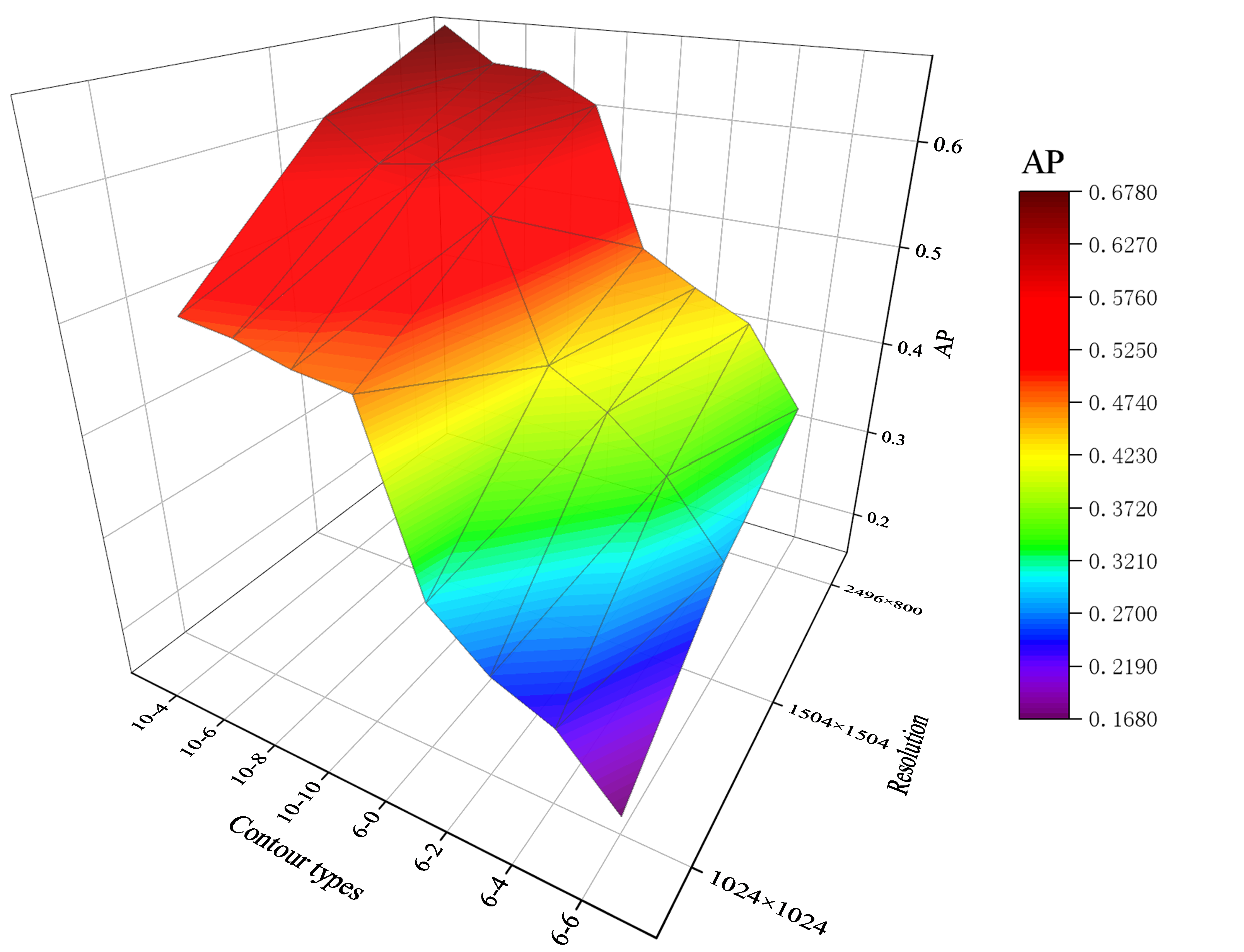}
	\caption{The relations among AP, resolution, and contour type: the highest AP is achieved at contour types is `10-4' and resolution is 2496$\times$800}
	\label{fig10}
\end{figure}

\textbf{Contour type}. The final results can be easily affected by the chosen contour type. The contour type experiment has been conducted to find the optimal one among contours mentioned in Fig. \ref{fig9}. We split these contour types into 4 groups in terms of the expansion of contours which are no expansion, 2 pixels expansion, 4 pixels expansion, and 6 pixels expansion. Each of them contains 2 types of annotation. `10-4' denotes the result of a 10 times dilation followed by a 4 times erosion. Only 6- and 10-times dilation are considered because we found less or more dilation operations had side effects including far too large annotation files and the overlap of annotated regions. Table \ref{tab3} compares the performance of different types of contours when the SOLOv2(Res-50-FPN) is utilized, and the size of the input image is 1024$times$1024. It can be seen from the data in Table \ref{tab3} that the `10-4' type gets the highest AP score, `the single most important metric'. Fig. \ref{fig10} illustrates AP values corresponding to various input scales and contour types. The best performance is achieved when the input is 2496$times$800 resolution using the `10-4' contour type.

\begin{table}[h]
	\centering
	\caption{INFLUENCE OF VARIOUS CONTOUR TYPES}
	\label{tab3}
	\begin{tabular}{ccccc}
		\hline\hline
		\begin{tabular}[c]{@{}c@{}}Expansion \\ Distance (px)\end{tabular} & Contour types & AP    & AP$_{50}$  & AP$_{75}$  \\\hline
		\multirow{2}{*}{0}      & 6-6           & 0.169 & 0.505 & 0.029 \\
		& 10-10         & 0.470 & \textbf{0.864} & 0.563 \\
		\multirow{2}{*}{2}      & 6-4           & 0.229 & 0.598 & 0.082 \\
		& 10-8          & 0.476 & 0.859 & 0.571 \\
		\multirow{2}{*}{4}      & 6-2           & 0.254 & 0.588 & 0.186 \\
		& 10-6          & 0.489 & 0.839 & 0.602 \\
		\multirow{2}{*}{6}      & 6-0           & 0.302 & 0.618 & 0.308 \\
		& 10-4          & \textbf{0.495} & 0.821 & \textbf{0.615}\\\hline\hline
	\end{tabular}
\end{table}

\subsection{Comparison Experiments}
The quantitative comparison experiments were conducted to compare our methods with other previous state-of-the-art approaches including Mask R-CNN and our improved version (Mask R-CNN*), Mask Scoring R-CNN \cite{39}, YOLACT \cite{40}, SOLO \cite{25}, and SOLOv2 \cite{7}. Configurations of these methods maintain as much consistency as possible. The batch size is set to 1 with the input image size of 2496$\times$800 and a `10-4' boundary type. Table \ref{tab4} provides the results of the above-mentioned methods on the PKLAD test set. Res-101-FPN denotes the backbone consisting of ResNet101 and FPN whereas the X-101-FPN represents using ResNeXt and HRFPN as the backbone. From Table \ref{tab4} we can see that these networks with the restricted size of the head, such as Mask R-CNN and YOLACT, achieve lower performance than the SOLOv2 which has no restriction on the mask head. The proposed method whose backbone consists of ResNeXt101 and HRFPN delivers the best performance across several metrics.

\begin{table}[h]
	\centering
	\caption{COMPARISON WITH PREVIOUS WORK}
	\label{tab4}
	\begin{tabular}{ccccc}
		\hline\hline
		Method      & Backbone    & AP    & AP$_{50}$  & AP$_{75}$  \\\hline
		Mask R-CNN  & Res-101-FPN & 0.247 & 0.568 & 0.134 \\
		Mask R-CNN* & Res-101-FPN & 0.260 & 0.580 & 0.161 \\
		MS R-CNN    & Res-101-FPN & 0.283 & 0.577 & 0.238 \\
		YOLACT      & Res-101-FPN & 0.232 & 0.544 & 0.157 \\
		SOLO        & Res-101-FPN & 0.331 & 0.730 & 0.172 \\
		SOLOv2      & Res-101-FPN & 0.706 & 0.913 & 0.820 \\
		Proposed    & X-101-HRFPN & \textbf{0.727} & \textbf{0.936} & \textbf{0.846}\\\hline\hline
	\end{tabular}
\end{table}

\begin{figure}[h]
	\centering
	\includegraphics[width=.6\columnwidth]{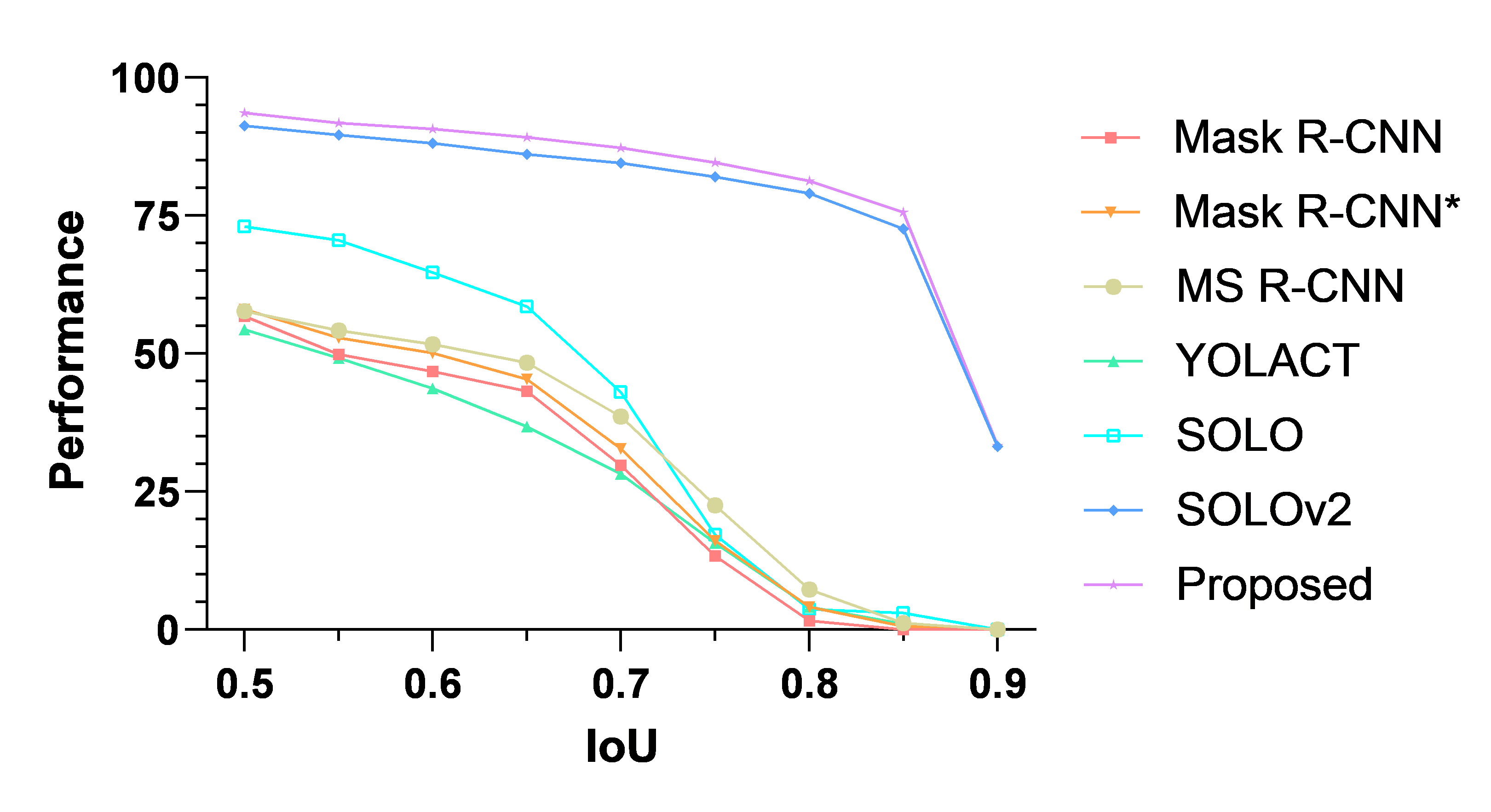}
	\caption{The performance of varied methods with different IoU thresholds}
	\label{fig11}
\end{figure}

APs of different methods with various IoU thresholds are illustrated in Fig. \ref{fig11}. It can be seen that the SOLOv2-based methods have no dramatic decrease until the IoU threshold reaches 85. Compared with Mask R-CNN or similar networks, the performance of SOLO is higher at the first beginning but experiences a sharp drop when the IoU threshold is over 65.

\begin{figure}[H]
	\centering
	\includegraphics[width=.6\columnwidth]{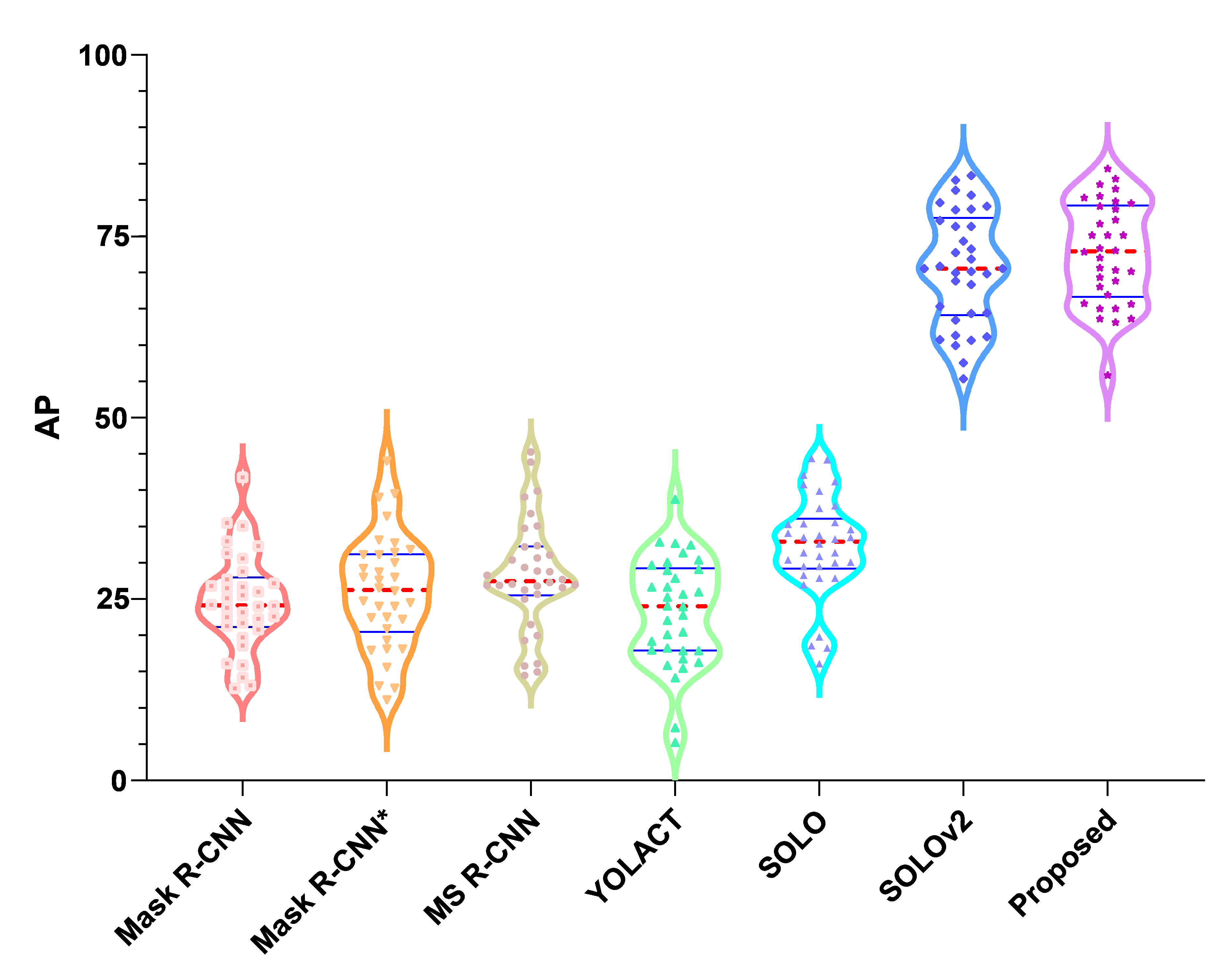}
	\caption{Violin plots of different methods}
	\label{fig12}
\end{figure}

The violin plots of different methods are exhibited in Fig. \ref{fig12}. The SOLOv2-based approaches obtained noticeably higher AP values. The AP values of results using our proposed method are more concentrated when comparing to the SOLOv2 which means that our methods may have stronger robustness.

Fig. \ref{fig13} provides a few visual examples of the fine-grained layout analysis result. Fig. \ref{fig13} (a) represents the sample with less sentence space, and the network is prone to treat a whole text line as an individual object. Fig. \ref{fig13} (b) is the visualization results of the sample with warped text lines. The sample with enough space between its sentences is shown in Fig. \ref{fig13} (c). Interestingly, the Latin letters, on the top of the image, have not been segmented into the text region incorrectly. Visual results of contaminated documents in Fig. \ref{fig13} (d) and (e) demonstrate the ability to deal with stains. Furthermore, it is noteworthy that the contour of the segmented text region sticks to the edges of the text lines which enables the precise segmentation of touching components between text lines.

\begin{figure}[H]
\centering
\vspace{-3mm}
	\subfigure[]{
		\begin{minipage}{.75\columnwidth}
			\includegraphics[width=\columnwidth]{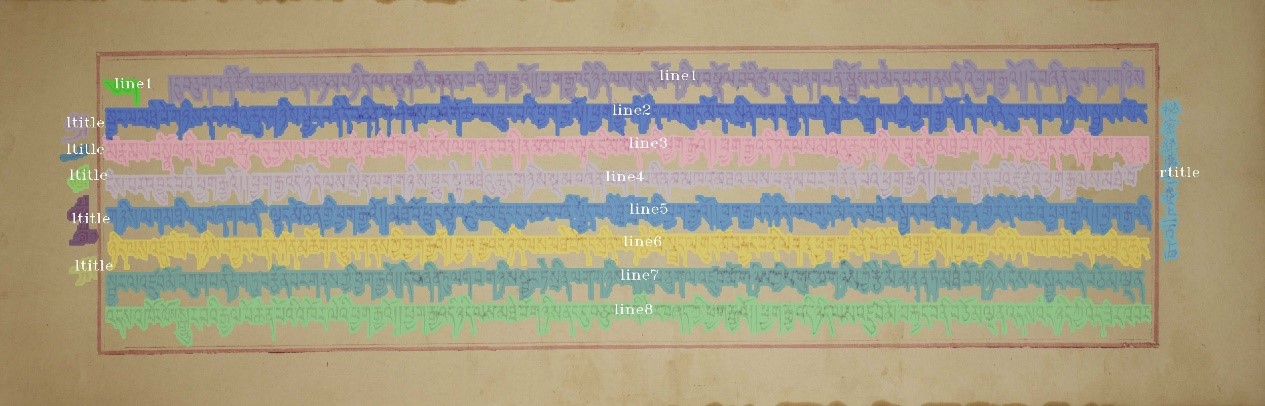}
	\end{minipage}}
\vspace{-3mm}

	\subfigure[]{
	\begin{minipage}{.75\columnwidth}
		\includegraphics[width=\columnwidth]{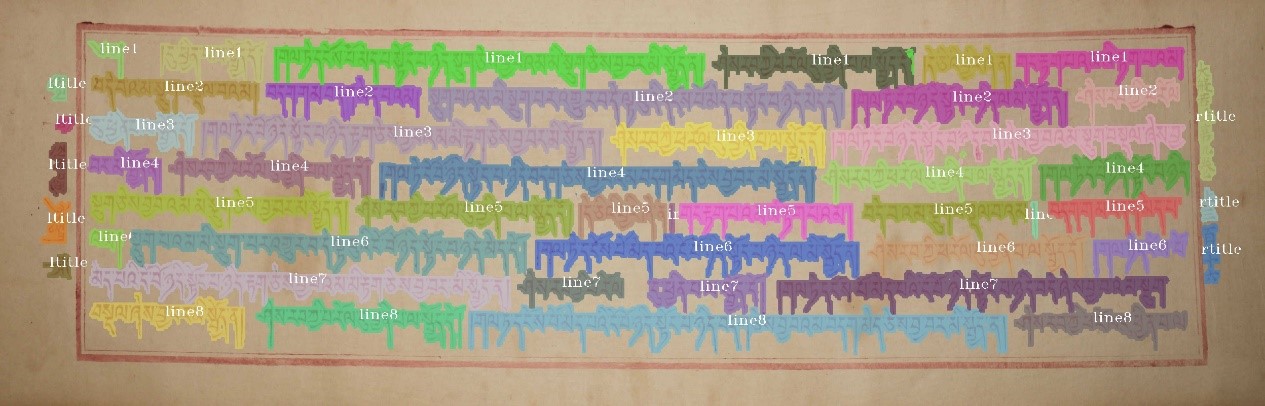}
\end{minipage}}
\vspace{-3mm}

	\subfigure[]{
	\begin{minipage}{.75\columnwidth}
		\includegraphics[width=\columnwidth]{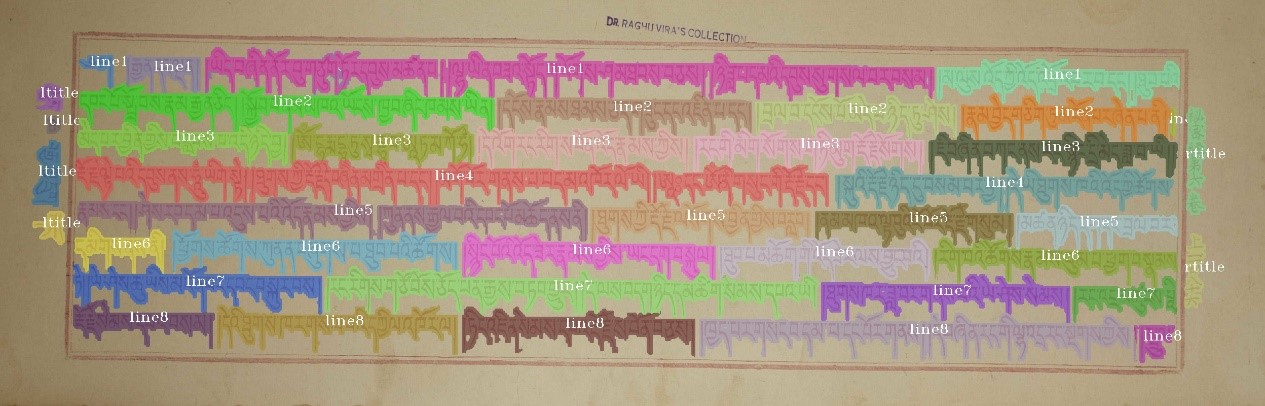}
\end{minipage}}
\vspace{-3mm}
\end{figure}
\begin{figure}[h]
\centering
	\subfigure[]{
	\begin{minipage}{.75\columnwidth}
		\includegraphics[width=\columnwidth]{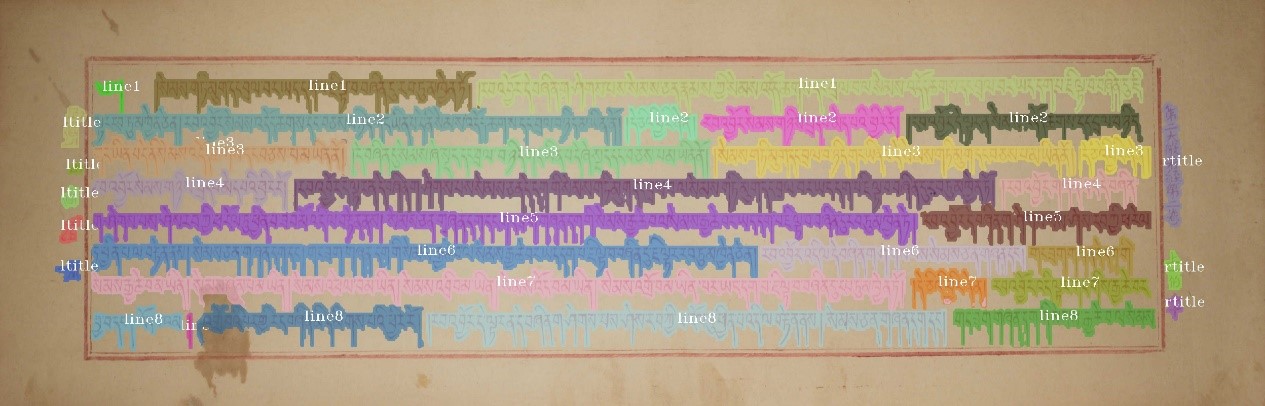}
\end{minipage}}
\vspace{-3mm}

	\subfigure[]{
	\begin{minipage}{.75\columnwidth}
		\includegraphics[width=\columnwidth]{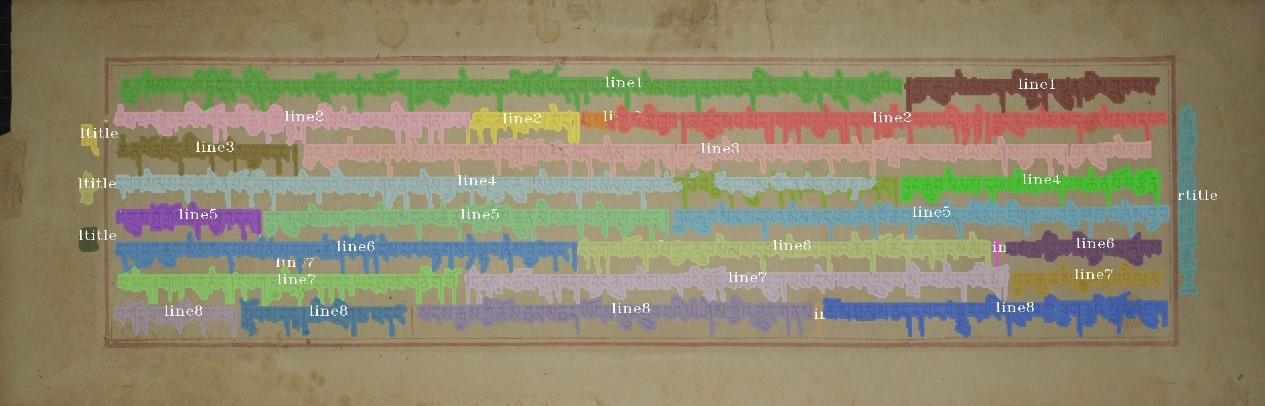}
\end{minipage}}

\caption{Visual Results: `linei' denotes instances that belong to the $i_{th}$ text line. `ltitle' and `rtitle' represent left and right titles respectively.}
\label{fig13}
\end{figure}

\subsection{Limitations}
\begin{figure}[h]
	\centering
	\vspace{-3mm}
	\subfigure[]{
		\begin{minipage}{.75\columnwidth}
			\includegraphics[width=\columnwidth]{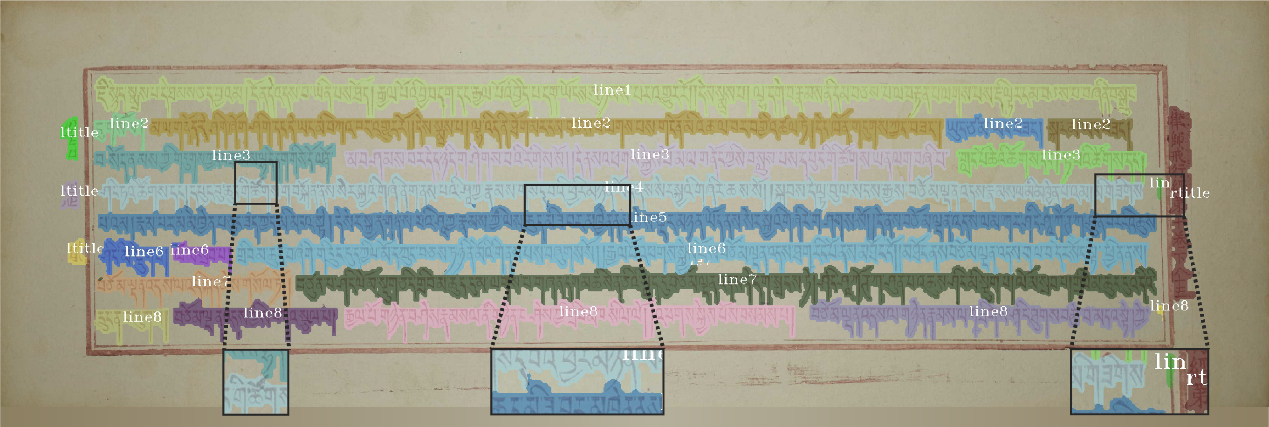}
	\end{minipage}}
	\vspace{-3mm}

	\subfigure[]{
		\begin{minipage}{.75\columnwidth}
			\includegraphics[width=\columnwidth]{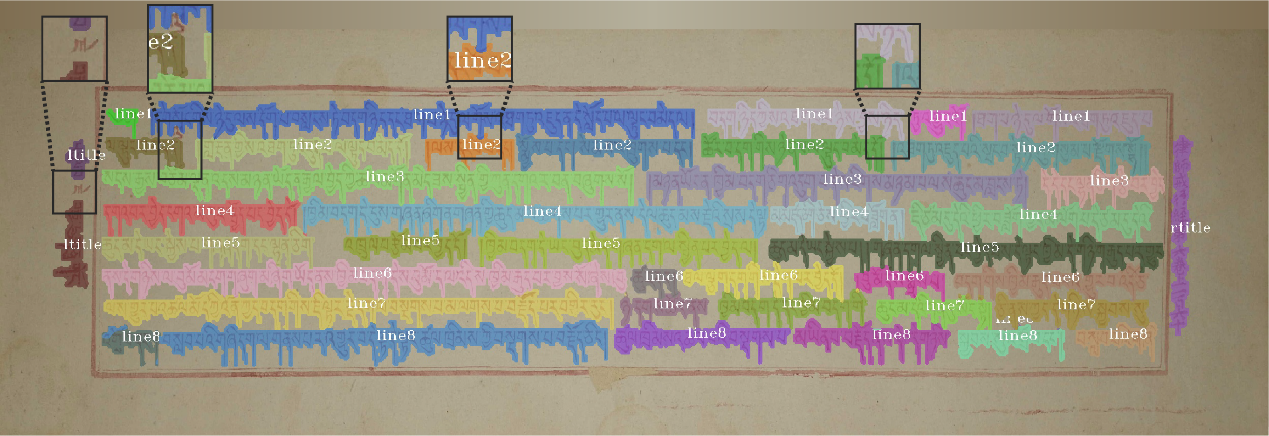}
	\end{minipage}}

	\caption{Failure cases include missing components and wrong segmentations.}
	\label{fig14}
\end{figure}

Though the proposed method produces decent results on most samples, defects can still be noticed in few visual results. Fig. \ref{fig1} (a) shows the cases where part of strokes or the borderline are allocated to the wrong text regions. In addition to wrong segmentation, failures like undetected characters and components (the two on the left) are exposed in Fig \ref{fig14}. (b). A possible explanation for this might be that the threshold of visualization of the masks discussed in Sec. 4.3 is a bit lower which is an inevitable trade-off between better visual effects and higher segmentation performance.

\section{Conclusions and future work}
This paper proposed parsing the layout of the historical Tibetan document images featuring non-uniform illumination, mottled background, various space between sentences, and considerable touching components. We treat layout analysis as an instance segmentation task and enhanced the SOLOv2 to identify different regions of the document. The experimental results indicate that our proposed end-to-end approach can perform accurate and fine-grained sub-line level layout analysis on the challenging historical Tibetan document images. Different text lines or sentence instances can be precisely separated and ready to be further recognized. The value of Average Precision reaches up to 72.7 when the input scale is 2496$\times$800 and the annotated contour is generated by a 10 times dilation and 4 times erosion. In addition, the proposed annotation method, which addresses the issues caused by the various space between sentences, accelerated the tedious process of labeling and helped to reduce human errors. Due to the lack of other public fine-grained datasets, we did not test our method on other document images. However, given that the fully convolutional network is language-agnostic, it is convinced that the proposed method can still deal with other documents in different languages with varied layouts.

Notwithstanding failures in some cases including wrong segmentation, the proposed method still rose to the most of challenges. A further study could focus on the model compression with the use of pruning and knowledge distillation to enable its deployment on the mobile device.

\bibliographystyle{unsrt}
\bibliography{ref}  






\end{document}